\documentclass[letterpaper, 10 pt, conference]{ieeeconf}  
\usepackage{graphicx}
 \usepackage{hyperref}

\IEEEoverridecommandlockouts                              

\usepackage{amsmath} 
\usepackage{amssymb}  
\usepackage{booktabs}
\usepackage{multirow}
\usepackage{hyperref}
\usepackage{threeparttable}

\usepackage{xcolor}
\usepackage{float}
\usepackage{caption}
\definecolor{darkgreen}{HTML}{32CD32}

\title{\LARGE \bf
MachaGrasp: Morphology-Aware Cross-Embodiment Dexterous Hand Articulation Generation for Grasping
}

\author{Heng Zhang$^{1,2}$, Kevin Yuchen Ma$^{1,3}$, Mike Zheng Shou$^{3}$, Weisi Lin$^{2}$ and Yan Wu$^{1*}$
\thanks{* denotes the corresponding author}
\thanks{$^{1}$Robotics \& Autonomous Systems Division, Institute for Infocomm Research, Agency for Science, Technology and Research (A*STAR-I$^{2}$R), Singapore {\tt\small wuy@i2r.a-star.edu.sg}}%
\thanks{$^{2}$College of Computing and Data Science, Nanyang Technological University, Singapore {\tt\small HENG018@e.ntu.edu.sg, wslin@ntu.edu.sg}}%
\thanks{$^{3}$Show Lab, National University of Singapore, Singapore {\tt\small yuchen\_ma@u.nus.edu, mikeshou@nus.edu.sg}}%
}

\begin{document}

\maketitle

\thispagestyle{empty}
\pagestyle{empty}

\begin{abstract}

Dexterous grasping with multi-fingered hands remains challenging due to high-dimensional articulations and the cost of optimization-based pipelines. Existing end-to-end methods require training on large-scale datasets for specific hands, limiting their ability to generalize across different embodiments. We propose MachaGrasp, an eigengrasp-based, end-to-end framework for \emph{cross-embodiment} grasp generation. From a hand’s morphology description, we derive a morphology embedding and an eigengrasp set. Conditioned on these, together with the object point cloud and wrist pose, an amplitude predictor regresses articulation coefficients in a low-dimensional space, which are decoded into full joint articulations. Articulation learning is supervised with a Kinematic-Aware Articulation Loss (KAL) that emphasizes fingertip-relevant motions and injects morphology-specific structure. In simulation on unseen objects across three dexterous hands, MachaGrasp attains a 91.9\% average grasp success rate with $\mathord{<}0.4\,\mathrm{s}$ inference per grasp. With few-shot adaptation to an unseen hand, it achieves 85.6\% success on unseen objects in simulation, and real-world experiments on this few-shot-generalized hand achieve an 87\% success rate.
The code and additional materials are available on  \href{https://connor-zh.github.io/MachaGrasp/}{our project website https://connor-zh.github.io/MachaGrasp/}.


\end{abstract}

\section{INTRODUCTION}

Dexterous grasping with multi-fingered robotic hands is a fundamental capability for versatile manipulation, offering rich contact interactions and adaptability to diverse object geometries with a spectrum of grasp solutions thanks to their inherent kinematic redundancy. However, the high-dimensional kinematics of such hands render grasp planning highly challenging.

Many existing methods are designed for a \emph{specific} hand~\cite{xu2024dexterous, wan2023unidexgrasp++, wang2025unigrasptransformer, zhang2025RobustDexGrasp}, requiring large-scale datasets and retraining whenever the embodiment changes.  This severely limits scalability, as each new hand design demands dedicated data collection and model training. Recent works have explored \emph{cross-embodiment} grasp generation, including DRO~\cite{11127754DRO}, DexGraspNet~\cite{wang2022dexgraspnet}, DFC~\cite{liu2021synthesizing}, UniGrasp~\cite{8972562}, and GenDexGrasp~\cite{li2023gendexgrasp}. These approaches either (i) directly optimize final hand poses via physics-based energy functions, or (ii) predict intermediate representations such as contact maps or robot–object distance matrices, which are then converted into an inverse-kinematics optimization problem. While effective, such optimization is often computationally expensive, particularly for complex hand morphologies.

\begin{figure}[!t]
    \centering
    \includegraphics[width=1\linewidth]{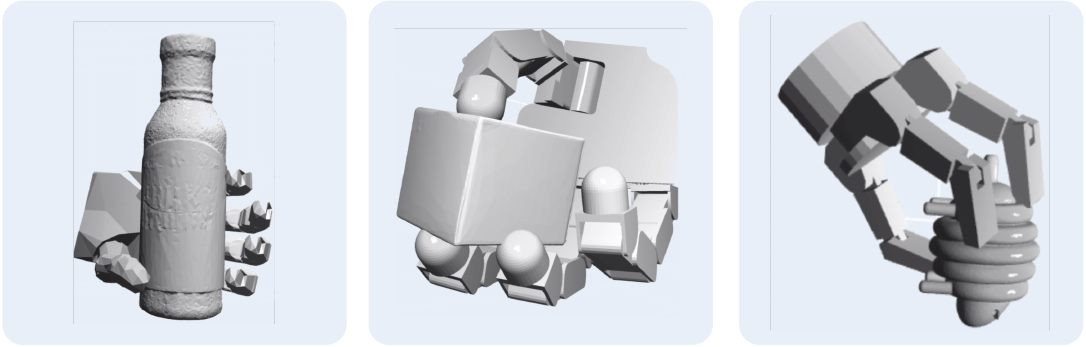}
    \caption{Visualization of generated grasps on test unseen objects. Left: ShadowHand, Middle: Allegro, Right: Barrett}
    \label{fig:visualization}
\end{figure}

End-to-end learning of articulations is attractive but challenging, as the dimensionality of hand degrees of freedom (DoFs) grows rapidly with morphological complexity. Santello et al.~\cite{santello1998postural} demonstrate that human grasp postures can be effectively represented in a low-dimensional space, with the first two principal components explaining over 80\% of the variance. Extending this outcome to robotics, Ciocarlie and Allen~\cite{ciocarlie2007dexterous} introduce \emph{eigengrasps}, hand-specific low-dimensional bases that capture coordinated joint patterns. Building on these findings, we hypothesize that a universal low-dimensional representation of grasp articulation could effectively reduce the search space for multi-finger grasp planning, facilitate end-to-end training, and support transfer of grasp skills across different robotic hand embodiments.

In this paper, we propose MachaGrasp, a framework for cross-embodiment dexterous grasp generation. MachaGrasp extracts eigengrasps and morphology embeddings directly from a robotic hand’s Unified Robot Description Format (URDF), explicitly incorporating kinematic and geometric constraints. Given an object point cloud and a target wrist pose, an amplitude predictor estimates articulation coefficients in the eigengrasp space, which are then decoded into full joint configurations. To supervise training, we introduce a \textbf{Kinematic-Aware Articulation Loss (KAL)} that emphasizes fingertip-relevant motions and implicitly encodes morphology-specific information, in contrast to naïve per-joint regression objectives (e.g., mean squared error) that overemphasize raw joint deviations.

We demonstrate the effectiveness of the proposed framework through extensive evaluations in both simulation and real-world settings. In simulation on unseen objects across three dexterous hands, MachaGrasp achieves a 91.9\% average grasp success rate with $<0.4\,\text{s}$ inference per grasp; with few-shot adaptation to an unseen hand it attains 85.6\% success on unseen objects in simulation, and real-world experiments on the same few-shot–generalized hand achieve 87\% success.

Our main contributions are:  
\begin{itemize}
    \item We propose MachaGrasp, an eigengrasp-based framework for \emph{cross-embodiment dexterous grasp generation} that predicts articulations in an end-to-end manner.
    \item We design a unified encoding scheme that converts a robot’s URDF into structured morphological tokens, capturing both kinematic constraints and geometric primitives.
    \item We propose a Kinematic-Aware Articulation Loss (KAL) that injects morphology-specific kinematic information into regression learning for articulation prediction, guiding the model beyond raw joint errors.
    \item We conduct extensive experiments in simulation and on real hardware, demonstrating effective grasping of novel objects across multiple robotic hands, including both seen hands and an unseen hand via few-shot adaptation.

\end{itemize}

\section{RELATED WORK}

\subsection{Learning-Based Dexterous Grasping}

Recent years have seen increasing interest in end-to-end methods for dexterous grasp generation, where neural networks directly map sensory inputs to grasp configurations. Dexterous Grasp Transformer \cite{xu2024dexterous} leverages large-scale datasets such as DexGraspNet \cite{wang2022dexgraspnet} to train a transformer-based model that predicts grasp configurations from object point clouds. Similarly, UniDexGrasp++ \cite{wan2023unidexgrasp++} and UniGraspTransformer \cite{wang2025unigrasptransformer} adopt unified frameworks to predict grasps for diverse objects, but their training and evaluation remain limited to a single robotic hand. Collectively, these works demonstrate the potential of scalable, end-to-end learning for dexterous grasping. However, their policies remain tied to the embodiments on which they were trained, limiting generalization across different hand morphologies.

\subsection{Cross-Embodiment Dexterous Grasping}

Another line of research focuses on enabling dexterous grasping across different robotic hands by leveraging hand-agnostic representations or shared action spaces. Cross-embodiment reinforcement learning approaches, such as the eigengrasp-based method by Yuan et al. \cite{yuan2025crossembodiment}, employ a shared low-dimensional action space to allow a single policy to operate across multiple hands, although training is often sample-intensive. 
More recently, GET-Zero \cite{patel2024getzero} introduces a graph-based embodiment transformer that conditions policy learning on the structural graph of the hand, achieving zero-shot transfer to unseen morphologies in dexterous in-hand manipulation tasks. While highly effective for manipulation, its focus is on policy learning rather than generating feasible grasps, leaving a gap that MachaGrasp addresses by focusing on effective grasp generation for diverse hand morphologies.

Liu et al. \cite{liu2021synthesizing} present an optimization-based approach that formulates a fast and differentiable force closure estimator, enabling the generation of diverse and physically stable grasps for arbitrary hand structures without relying on training data. While highly general, such optimization-based approaches \cite{liu2021synthesizing, fastd,turpin2022grasp} can be computationally intensive and are subject to careful tuning for efficient execution.

Another direction is to learn intermediate hand-agnostic representations followed by optimization. GenDexGrasp \cite{li2023gendexgrasp} predicts dense contact maps on the object surface, which are then used to recover hand-specific grasps through iterative optimization. Attarian et al. \cite{attarian2023geometry} similarly aligns objects and hands in a shared metric space to generate transferable grasps. DexRepNet \cite{liu2023dexrepnet}, AnyDexGrasp \cite{fang2025anydexgrasp}, and UniGrasp \cite{8972562} leverage object-centric or contact-centric latent representations to enable cross-hand generalization, while DRO \cite{11127754DRO} encodes the interaction between robot and object as a distance matrix between point clouds, from which grasps are recovered using multilateration. These approaches demonstrate promising cross-embodiment performance, but iterative optimization or per-hand modules often limit scalability, efficiency, or training stability.

\begin{figure*}[t!]
    \centering
    \includegraphics[width=1\textwidth]{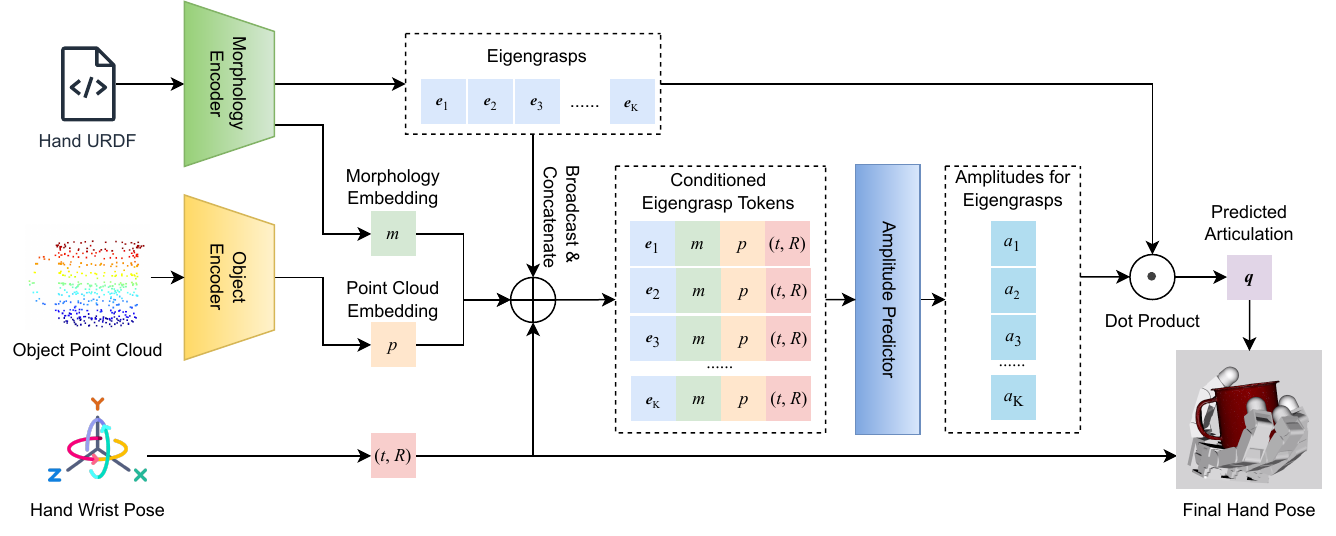}
    \caption{Method Overview. MachaGrasp processes hand URDF, object point cloud, and wrist pose through specialized encoders to generate morphology embedding $\boldsymbol{m}$, point cloud embedding $\boldsymbol{p}$, and pose encoding $\boldsymbol{h}$. These are fused to predict eigengrasp amplitudes $\boldsymbol{a}$, which are combined with eigengrasps $E$ to produce the final articulation vector $\boldsymbol{q}$.}

    \label{fig:architecture}
\end{figure*}

\section{METHOD}

\subsection{Problem Formulation}

We formulate dexterous hand articulation generation as the problem of predicting joint configurations conditioned on the geometry of the target object, the morphology of the hand, and the pose of the wrist.

\textbf{Target Object Representation.} The target object is represented by a point cloud
$P \in \mathbb{R}^{N\times3}$ ,
where $N$ is the number of sampled points. For consistency across objects, the point cloud is normalized by translating its centroid to the origin of the world frame.

\textbf{Hand Morphology.} Each robotic hand is described by its URDF specification, from which we extract a set of joint encodings
$J = \{\boldsymbol{j}_k\}_{k=1}^M $,
where $M$ denotes the number of joints. Each encoding $j_k$ captures the structural and kinematic properties of the joint and its associated links, providing a unified representation of morphology.

\textbf{Wrist Pose.} The global placement of the hand is defined by the wrist transformation $t \in \mathbb{R}^3$ and orientation $R \in SO(3)$, both expressed in the world frame. In implementation, we represent $R$ using the continuous 6D rotation parameterization~\cite{Zhou_2019_CVPR} for stability.

\textbf{Articulation Output.} The goal is to predict the articulation vector $q \in \mathbb{R}^d$, where $d$ depends on the number of joints of the given hand. To enable consistency across different morphologies, we adopt an eigengrasp formulation in which the articulation is expressed as a linear combination of $K$ basis vectors $\{\boldsymbol{e}_i\}_{i=1}^K$:
\begin{equation}
\boldsymbol{q} = \sum_{i=1}^K a_i \boldsymbol{e}_i ,
\label{q=ae}
\end{equation}
where $a_i \in \mathbb{R}$ are the predicted amplitudes. In our implementation, we set $K=9$.

\subsection{Method Overview}

An overview of MachaGrasp is shown in Fig.~\ref{fig:architecture}. The framework takes as input the hand URDF, the object point cloud, and the wrist pose. The hand URDF is first processed by the morphology encoder (Sec.~\ref{sec:Morphology Encoder}), which produces a morphology embedding ($\boldsymbol{m}$) together with a set of eigengrasps ($E$). The object point cloud is then processed by the object encoder (Sec.~\ref{sec:Object Encoder}) to extract a point cloud embedding ($\boldsymbol{p}$). These embeddings, combined with the wrist pose ($\boldsymbol{h}$), are fused to form conditioned eigengrasp tokens. The amplitude predictor (Sec.~\ref{sec:Amplitude Predictor}) takes these tokens as input and estimates the amplitudes for each eigengrasp ($a$). Finally, the articulation vector ($\boldsymbol{q}$) is obtained by the dot product between the predicted amplitudes and the eigengrasps, and together with the wrist pose, defines the final hand pose.


\subsection{Morphology Encoder}
\label{sec:Morphology Encoder}

\begin{figure}[t]
    \centering
    \includegraphics[width=0.9\linewidth]{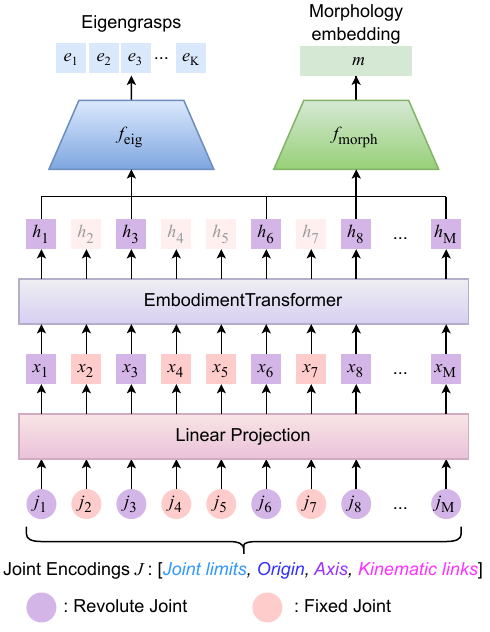}
    \caption{Architecture of Morphology Encoder: The joint encodings are mapped into tokens and then processed by the EmbodimentTransformer. Relevant output tokens corresponding to revolute joints are concatenated and used by the morphology head and eigengrasp heads to produce the morphology embedding and eigengrasps respectively.}
    \label{fig:morph_encoder}
\end{figure}

A central component of MachaGrasp is the morphology encoder (illustrated in Fig. \ref{fig:morph_encoder}), which learns structured representations of dexterous hands directly from their URDF descriptions. In contrast to prior work that encodes hand morphology solely as mesh point clouds~\cite{8972562,liu2021synthesizing,wang2022dexgraspnet}, MachaGrasp leverages the URDF to extract explicit kinematic constraints and geometric primitives. Based on this structured information, the encoder predicts both the eigengrasps of the hand and a compact latent representation that captures its underlying morphology.

\textbf{URDF-based joint encodings.}  
From the URDF of each hand, we construct a set of joint encodings $J$.
Each joint $\boldsymbol{j}_k$ includes:  
\begin{itemize}
    \item \textit{Joint limits:} minimum and maximum values (radians).  
    \item \textit{Origin:} pose in the parent frame, parameterized by roll, pitch, yaw, and Cartesian translation $(r, p, \psi, x, y, z)$.  
    \item \textit{Axis:} unit vector specifying the motion direction.  
    \item \textit{Kinematic links:} parent and child links, each approximated by a primitive shape with pose and size parameters. We adopt a uniform encoding:  
    \begin{align}
    \text{Box: } & (0, r, p, \psi, x, y, z, \text{length}, \text{width}, \text{height}), \nonumber \\
    \text{Cylinder: } & (1, r, p, \psi, x, y, z, \text{length}, \text{radius}, 0), \nonumber \\
    \text{Sphere: } & (2, r, p, \psi, x, y, z, \text{radius}, 0, 0), \nonumber \\
    \text{Dummy: } & (3, 0, 0, 0, 0, 0, 0, 0, 0, 0), \nonumber
    \end{align}
    where the leading integer denotes the primitive type (0: box, 1: cylinder, 2: sphere, 3: dummy). The dummy type provides compatibility for non-physical links without geometry.
\end{itemize}

\textbf{Tokenization and embedding.}  
Each joint encoding $\boldsymbol{j}_k$ is mapped into a feature vector using MLP embeddings, with joint and link features sharing the same embedding layers. The resulting sequence of tokens is concatenated into 
$X \in \mathbb{R}^{M \times d_h}$,  
and normalized to balance feature scales.  

\textbf{EmbodimentTransformer with masking.}  
To model structural dependencies among joints, we employ the EmbodimentTransformer from GET-Zero~\cite{patel2024getzero}, which is designed for zero-shot embodiment generalization across diverse morphologies. It treats each joint and link as a token and applies self-attention to capture kinematic relationships. Given the token sequence $X$, the encoder produces morphology-aware features while ignoring padded tokens through a source key padding mask:  
\begin{equation}
H = \text{EmbodimentTransformer}(X;\,\text{mask}(M)).
\end{equation}
Here $H \in \mathbb{R}^{M \times d_h}$ denotes the morphology-aware feature sequence.  

\textbf{Revolute-joint selection.}  
Since only revolute joints contribute to articulation, their features are extracted using a binary mask $\boldsymbol{\rho} \in \{0,1\}^M$. The resulting variable-length sequence is padded to a fixed length $D_{\max}$, yielding  
$
H' \in \mathbb{R}^{D_{\max} \times d_h},
$  
together with an associated padding mask for downstream processing.  

\textbf{Outputs.}  
Two prediction heads operate on $H'$. The \emph{morphology head} applies an MLP with attention pooling to produce a compact embedding  
\begin{equation}
\boldsymbol{m} = f_{\text{morph}}(H'), \qquad \boldsymbol{m} \in \mathbb{R}^{d_m},
\end{equation}
which summarizes the kinematic and geometric structure of the hand. In parallel, $K$ \emph{eigengrasp heads} predict low-dimensional basis vectors for articulation:  
\begin{equation}
E = \{\boldsymbol{e}_i\}_{i=1}^{K} = f_{\text{eig}}(H'), \qquad E \in \mathbb{R}^{K \times D_{\max}},
\end{equation}
with masked positions zeroed and $E$ interpreted in the true DoF space via $\boldsymbol{\rho}$.  

In summary, the morphology encoder transforms URDF-derived representations into a compact morphology embedding $m$ and a set of eigengrasps $E$, jointly providing an expressive description of hand embodiment for conditioning articulation prediction.

\subsection{Object Encoder}
\label{sec:Object Encoder}

The object encoder extracts geometric features from the target object point cloud, which serves as a condition for articulation prediction.
We employ a hierarchical PointNet++ backbone~\cite{qi2017pointnetplusplus} to process the point cloud.
Specifically, the encoder consists of three set abstraction (SA) modules:
\begin{itemize}
    \item The first SA module samples 128 points within a radius of 0.02 and applies multi-layer perceptrons (MLPs) with dimensions $[64,64,128]$ to extract local geometric features.  
    \item The second SA module samples 32 points within a radius of 0.04 and applies MLPs with dimensions $[128,128,256]$ to capture higher-level shape information.  
    \item The final SA module aggregates all points globally and applies MLPs with dimensions $[256, 512, 1024]$ to produce a global object representation.
\end{itemize}
The output of the encoder is a compact embedding representing the global geometry of the object:
\begin{equation}
\boldsymbol{f}_{\text{obj}} = \text{PointNet++}(P), \qquad f_{\text{obj}} \in \mathbb{R}^{1024}
\end{equation}

To improve the quality of the learned representation, the object encoder is pretrained as part of an autoencoder. Specifically, a PointNet++ encoder and a point cloud decoder are jointly trained to minimize the Chamfer distance \cite{Fan_2017_CVPR} between the input point cloud and its reconstruction:
\begin{equation}
\mathcal{L}_{\text{CD}} = d_{\text{Chamfer}}(\hat{P}, P),
\end{equation}
where $\hat{P}$ is the reconstructed point cloud. The pretrained encoder is then integrated into the full model.

\subsection{Amplitude Predictor}
\label{sec:Amplitude Predictor}
The amplitude predictor estimates the coefficients associated with each eigengrasp, conditioned on the object, hand morphology, and wrist pose. Given the eigengrasps $E = \{\boldsymbol{e}_i\}_{i=1}^K$, the morphology embedding $m$, the object feature $f_{\text{obj}}$, and the wrist pose $(t, R)$, the predictor outputs amplitudes $\boldsymbol{a} = \{a_i\}_{i=1}^K$ that define the articulation.  

For each eigengrasp $\boldsymbol{e}_i$, we construct a \textit{Conditioned Eigengrasp Token} by concatenating its basis vector with the replicated morphology, object, and wrist pose encodings:
\begin{equation}
\boldsymbol{\tau}_i = [\, \boldsymbol{e}_i,\, \boldsymbol{m},\, \boldsymbol{f}_{\text{obj}},\, t,\, R \,],
\end{equation}
where $\boldsymbol{\tau}_i \in \mathbb{R}^{d_\tau}$. The resulting sequence of $K$ Conditioned Eigengrasp Tokens, $\{\boldsymbol{\tau}_i\}_{i=1}^K$, is normalized, projected into the transformer input space, and processed by a transformer encoder:
\begin{equation}
Z = \text{TransformerEncoder}(\{\tau_i\}_{i=1}^K),
\end{equation}
where $Z \in \mathbb{R}^{K \times d_h}$ is the context-aware eigengrasp feature sequence.  

Each feature $Z_i$ is then passed through an amplitude head specific to eigengrasp $\boldsymbol{e}_i$:
\begin{equation}
a_i = f_{\text{amp},i}(Z_i),
\end{equation}
yielding the final amplitude set $\boldsymbol{a}$.  

The predicted amplitudes are combined with the eigengrasps to reconstruct the articulation vector following Eq.~\ref{q=ae}.
This design enables the predictor to model interactions among object geometry, hand morphology, and wrist pose, while assigning a dedicated prediction pathway to each eigengrasp.

\begin{table*}[t]
\centering
\caption{Comparison on 28 unseen objects using \emph{predicted} wrist poses from 6-DoF GraspNet (50 candidates/object).}
\label{tab:pred_wrist_pose}
\resizebox{0.9\textwidth}{!}{
\begin{tabular}{l|cccc|cccc}
\toprule
\multirow{2}{*}{Method} 
& \multicolumn{4}{c|}{Success Rate (\%) $\uparrow$} 
& \multicolumn{4}{c}{Efficiency (sec.) $\downarrow$} \\
& Avg. & ShadowHand & Allegro & Barrett 
& Avg. & ShadowHand & Allegro & Barrett \\
\midrule
GraspIt!~\cite{graspit} & 89.9 & $-^{\star}$ & \textbf{92.0} & 87.7 
& $-$ & $-$ & $-$ & $-$ \\
DexGraspNet~\cite{wang2022dexgraspnet} & 64.1 & 86.1 & 45.4 & 60.9 
& $>480$ & $>800$ & $>360$ & $>260$ \\
DRO~\cite{11127754DRO} & 89.1 & 80.0 & 90.7 & \textbf{96.5} 
& 0.564 & 0.801 & 0.459 & 0.432 \\
\midrule
MachaGrasp (w/ MSE) & 90.2 & 88.4 & 90.6 & 91.6
& 0.357 & 0.359 & \textbf{0.350} & 0.361 \\
\textbf{MachaGrasp (w/ KAL)} & \textbf{91.9} & \textbf{90.7} & 91.8 & 93.1 
& \textbf{0.353} & \textbf{0.354} & 0.351 & \textbf{0.353} \\
\bottomrule
\end{tabular}
}
\begin{tablenotes}
\small
\item $^{\star}$ ShadowHand result for GraspIt! is not reported due to severe self-collision issues in the generated grasps.
\end{tablenotes}
\end{table*}

\subsection{Loss Function}

The training objective combines two complementary terms: an eigengrasp regression loss and a kinematic-aware articulation loss (KAL).  

\textbf{Eigengrasp loss.} The morphology encoder predicts a set of eigengrasps $E = \{e_i\}_{i=1}^K$, which are supervised against ground-truth eigengrasps $E^{*} = \{e_i^{*}\}_{i=1}^K$ obtained via principal component analysis (PCA) on training articulation data. This alignment is enforced using a mean squared error (MSE) loss:
\begin{equation}
\mathcal{L}_{\text{eig}} = \frac{1}{K} \sum_{i=1}^K \| e_i - e_i^{*} \|_2^2.
\end{equation}

\textbf{Kinematic-Aware Articulation Loss (KAL).} A uniform MSE loss on joint values does not account for the fact that different joints contribute unequally to fingertip motion. Proximal joints typically induce larger fingertip displacements due to longer lever arms, while distal joints provide fine-grained local adjustments. To capture this asymmetry, we employ a Jacobian-guided weighting scheme. For each hand embodiment, fingertip Jacobians are computed with respect to the ground-truth articulation $q^{*}$, and per-joint weights are derived from the squared, weighted Jacobian entries:
\begin{equation}
w_f = \sum_{r=1}^6 \lambda_r J_{r,:}^2 ,
\end{equation}
where $J \in \mathbb{R}^{6 \times d_f}$ is the Jacobian of a finger with $d_f$ joints, and $\lambda_r$ assigns higher importance to translational components ($\lambda_{1:3}=1.0$) and lower importance to rotational components ($\lambda_{4:6}=0.05$). The weights are normalized such that their mean is approximately one.  

The articulation loss is then defined as:
\begin{equation}
\mathcal{L}_{\text{KAL}} = \frac{1}{d} \sum_{j=1}^d w_j \, (q_j - q_j^{*})^2 ,
\end{equation}
where $w_j$ is the Jacobian-derived weight for joint $j$, and $q_j$, $q_j^{*}$ are the predicted and ground-truth joint values.  

\textbf{Final objective.} The overall training loss is the sum of the two terms:
\begin{equation}
\mathcal{L} = \mathcal{L}_{\text{eig}} + \mathcal{L}_{\text{KAL}} .
\end{equation}

This formulation enforces eigengrasp consistency and articulation accuracy, while KAL guides the model to respect the functional kinematics of the hand rather than merely minimizing raw numeric joint errors. Since the Jacobian weighting is computed at the ground-truth articulation and depends on the embodiment’s kinematic structure, the loss implicitly encodes morphology-specific information, providing task-aware supervision that emphasizes fingertip-relevant motions and supports generalization across hands with different designs.

\subsection{Domain Randomization and Data Augmentation}

To improve robustness and generalization, we apply domain randomization and data augmentation during training.  

\textbf{Geometric randomization.} The object point cloud and the corresponding wrist pose are jointly rotated in the world frame by a random Euler rotation. Each rotation angle is sampled uniformly from $[-20^\circ, 20^\circ]$, ensuring consistency between object geometry and hand placement while exposing the model to diverse global configurations.  

\textbf{Noise injection.} We further perturb the training data with Gaussian noise:  
\begin{itemize}
    \item Point cloud coordinates are corrupted with zero-mean Gaussian noise scaled by $\sigma_{\text{pcl}} = 0.002$.  
    \item Wrist translation and wrist orientation are perturbed with noise scaled by $\sigma_{\text{trans}} = 0.001$ and $\sigma_{\text{rot}} = 0.01$, respectively.  
    \item Articulation vectors are perturbed with Gaussian noise scaled by $\sigma_{\text{art}} = 0.002$.  
\end{itemize}

These augmentations encourage the model to focus on robust feature extraction rather than overfitting to exact point cloud coordinates or pose parameters, thereby improving transferability across different embodiments and object instances.

\section{EXPERIMENTS}

Our experiments are designed to answer the following key questions:

\begin{itemize}
    \item \textbf{Q1:} How well does MachaGrasp perform in realistic settings where wrist poses are predicted by a naive pose predictor, compared to state-of-the-art baselines?
    \item \textbf{Q2:} Does integrating articulation predictions from MachaGrasp with the wrist poses generated by existing baselines (GraspIt, DexGraspNet, DRO) improve their grasp success rates?
    \item \textbf{Q3:} Can MachaGrasp generalize to previously unseen robotic hands with only few-shot adaptation?
    \item \textbf{Q4:} Does the learned policy transfer effectively from simulation to real-world hardware?
\end{itemize}

\begin{figure*}[t]
    \centering
    \includegraphics[width=0.9\linewidth]{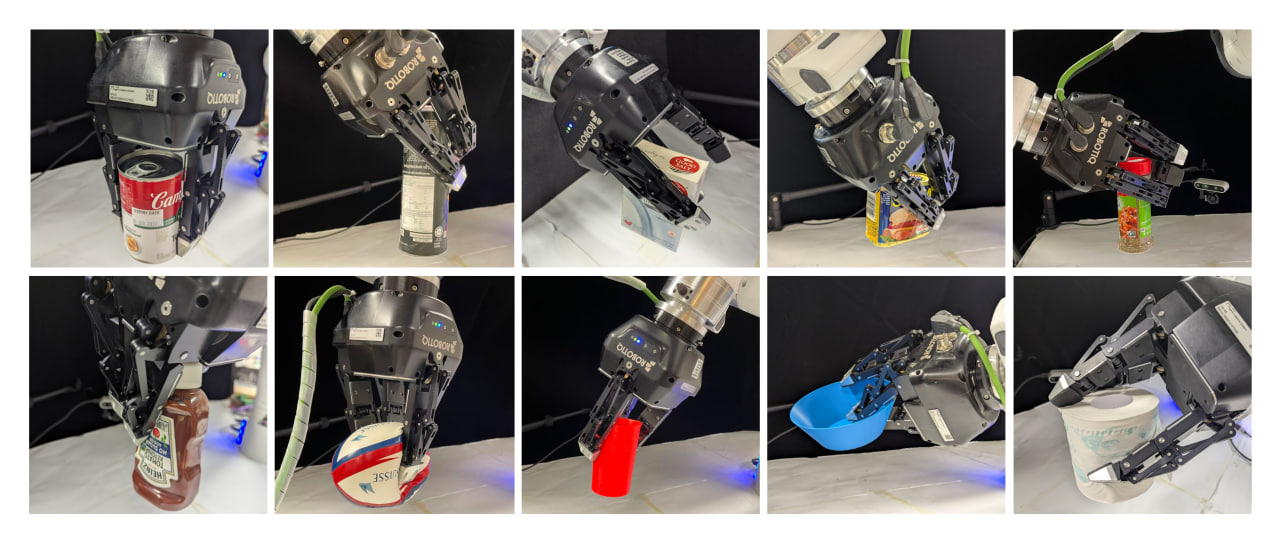}
    \caption{Real-world experiment test objects and example predicted grasps.}
    \label{fig:real_exp}
\end{figure*}

\subsection{Datasets}
\label{sec:exp-dataset}

We train MachaGrasp on three dexterous robotic hands: ShadowHand (5 fingers, 22 DoF), Allegro Hand (4 fingers, 16 DoF), and Barrett Hand (3 fingers, 8 DoF). For Allegro and Barrett, we adopt grasp data from the MultiGripperGrasp dataset~\cite{casas2024multigrippergrasp}. Since the ShadowHand subset in MultiGripperGrasp suffers from severe self-collision artifacts, we instead use DexGraspNet~\cite{wang2022dexgraspnet} for ShadowHand data.  

To ensure grasp stability, all datasets are filtered in simulation using Isaac Gym~\cite{liang2018gpu} (see Sec.~\ref{sec:isaac-sim-setup}), and only stable grasps are retained. After filtering, we construct splits at both the \emph{object} and \emph{pose} levels. A subset of objects is entirely withheld for unseen-object evaluation, while the remaining objects are divided into training, validation, and seen-test splits following an 8:1:1 ratio.  

In total, the filtered dataset contains approximately 1.69M stable grasps across 6,029 unique objects.

\subsection{Simulation Setup}
\label{sec:isaac-sim-setup}

Building on the evaluation protocol of DexGraspNet~\cite{wang2022dexgraspnet}, we use Isaac Gym~\cite{liang2018gpu} with PhysX as the underlying physics engine to verify the stability of candidate grasps. In each trial, the object is initialized in a floating state within the scene, and the hand is set to the target pose (translation, orientation, and articulation). Gravity is then enabled, allowing the object to settle naturally under the grasp configuration. A grasp is labeled as stable if the object remains securely in contact with the hand for 200 simulation steps without slipping or falling.  

\subsection{Articulation Prediction with Predicted Wrist Pose}
\label{sec:exp-graspnet}

To answer Q1, we evaluate MachaGrasp in a realistic setting where wrist poses are \emph{predicted} rather than provided. Specifically, we use 6-DoF GraspNet~\cite{mousavian2019graspnet} to generate parallel-gripper candidates per object, and map each candidate to a hand-specific wrist pose using a calibrated rigid transform $\Delta_{\text{hand}}$:
\[
T_{\text{wrist}}^{(\text{hand})} = \Delta_{\text{hand}} \, T_{\text{gripper}}^{(\text{GraspNet})}.
\]
Since some GraspNet proposals are of low quality or collide with the target object, we adopt the learned grasp stability evaluator from~\cite{taunyazov2023refining} to filter out infeasible poses before articulation prediction.  

For each retained wrist pose $T_{\text{wrist}}^{(\text{hand})}$, MachaGrasp predicts an articulation, and the resulting grasp is evaluated in simulation. Experiments are conducted on a test set of \textbf{28 unseen objects}, with \textbf{50} candidate grasps per object. Example grasps generated by MachaGrasp are visualized in Fig.~\ref{fig:visualization}.

We compare against three baselines:  
\begin{itemize}

    \item \textbf{GraspIt!}~\cite{graspit}: Generates grasps by closing hand joints at preset velocities until contact or joint limits are reached, followed by force-closure analysis. We evaluate using grasps from the MultiGripperGrasp dataset originally generated by GraspIt! and filtered through its simulation protocol. As these grasps underwent physics-based filtering during dataset creation, the reported success rates likely reflect an optimistic estimate of baseline performance. We do not evaluate its ShadowHand result due to strong self-collision issues in the dataset (Sec.~\ref{sec:exp-dataset}).

    \item \textbf{DexGraspNet}~\cite{wang2022dexgraspnet}: Generates grasp poses by sampling candidates in simulation, then refining them with the differentiable force-closure estimator in DFC~\cite{liu2021synthesizing} and physics-based optimization. While the original implementation supports only the ShadowHand, we re-implemented the pipeline for the Allegro and Barrett Hands.

    \item \textbf{DRO}~\cite{11127754DRO}: The state-of-the-art in cross-embodiment grasp generation. It predicts a distance-based hand–object interaction representation and recovers the final pose via optimization. For fairness, we retrained DRO from scratch on our dataset until convergence.  
\end{itemize}

As an ablation, we also compare MachaGrasp trained with the proposed \textbf{KAL loss} against the standard MSE loss. All methods are evaluated on the same object set with an equal number of candidates. Following the evaluation protocol of DRO, we report two complementary metrics:
\begin{itemize}
    \item \textbf{Success rate:} fraction of stable grasps under the simulation protocol.  
    \item \textbf{Efficiency:} average computation time per successful grasp, including network inference and post-processing.
\end{itemize}

\noindent\textbf{Result Analysis.}
Table~\ref{tab:pred_wrist_pose} summarizes performance across ShadowHand, Allegro, and Barrett. MachaGrasp achieves the highest overall success rate (91.9\%) while also being the most efficient ($0.353\,\text{s}$). Compared to DRO, we obtain substantial gains on ShadowHand (+10.7\%) and Allegro (+1.1\%). The only exception is the Barrett hand, where DRO obtains a slightly higher success rate. This can be attributed to DRO's optimization-based nature, which is more effective on hands with lower degrees of freedom, as also suggested by the trend in Table~\ref{tab:pred_wrist_pose}. 

More notably, MachaGrasp outperforms DexGraspNet across all hand types and achieves significantly better efficiency. We found DexGraspNet highly sensitive to hyperparameter settings across different embodiments. Despite extensive tuning, performance varied, highlighting a key drawback of optimization-based methods-reliance on hand-specific hyperparameter tuning. In contrast, MachaGrasp maintains consistently high performance across all hand types without requiring hand-specific parameter adjustment.

\noindent\textbf{Ablation.}
Comparing loss functions, the proposed KAL loss further boosts performance over MSE, improving the average success rate by 1.7\%. This confirms the benefit of KAL in learning articulation representations that transfer robustly across different robotic hands.

\noindent\textbf{Efficiency.}
Since 6-DoF GraspNet cannot be run on newer GPUs due to software dependencies, all timings are measured on an NVIDIA RTX 2080 Ti GPU. Our end-to-end method is expected to gain further efficiency improvements on higher-end hardware.

\begin{table}[h]
\centering
\caption{Grasp success rates and improvement (\%) using wrist poses from baseline methods with MachaGrasp.}
\label{tab:baseline_wrist_pose}
\resizebox{0.49\textwidth}{!}{
\begin{tabular}{lcccc}
\toprule
\parbox{1.5cm}{\raggedright Wrist Pose Source} & Average & ShadowHand & Allegro & Barrett \\
\midrule
GraspIt!~\cite{graspit}        & 92.4(\textcolor{darkgreen}{+2.5}) & $-^\star$ & 93.5(\textcolor{darkgreen}{+1.5}) & 91.6(\textcolor{darkgreen}{+3.9}) \\
DexGraspNet~\cite{wang2022dexgraspnet} & 68.8(\textcolor{darkgreen}{+4.7}) & 90.7(\textcolor{darkgreen}{+4.6}) & 52.1(\textcolor{darkgreen}{+6.7}) & 63.7(\textcolor{darkgreen}{+2.8}) \\
DRO~\cite{11127754DRO}          & 93.6 (\textcolor{darkgreen}{+4.5}) & 94.7 (\textcolor{darkgreen}{+14.7}) & 96.0 (\textcolor{darkgreen}{+5.3}) & 90.2 (\textcolor{red}{-6.3}) \\
\bottomrule
\end{tabular}
}
\begin{tablenotes}
\small
\item $^{\star}$ ShadowHand result for GraspIt! is not reported due to severe self-collision issues in the generated grasps.
\end{tablenotes}
\end{table}

\vspace{-5pt}

\subsection{Articulation Prediction with Baseline Wrist Poses}

We further evaluate MachaGrasp on the unseen-object test set using wrist poses generated by existing baselines (Sec.~\ref{sec:exp-graspnet}). 
Specifically, we take the wrist poses proposed by DFC~\cite{liu2021synthesizing}, DexGraspNet~\cite{wang2022dexgraspnet}, and DRO~\cite{11127754DRO}, and apply MachaGrasp to generate the corresponding articulations. 
This experiment examines whether the predicted articulations can improve grasp success when paired with wrist poses from external methods (Q2).  

Table~\ref{tab:baseline_wrist_pose} summarizes the grasp success rates. The numbers in brackets indicate the improvement in success rate compared to the baseline methods. The results indicate that MachaGrasp demonstrates consistent improvements when applied to baseline wrist poses across most configurations, suggesting the potential for greater performance increase when paired with higher quality wrist pose generation.

\begin{figure}[t]
    \centering
    \includegraphics[width=0.85\linewidth]{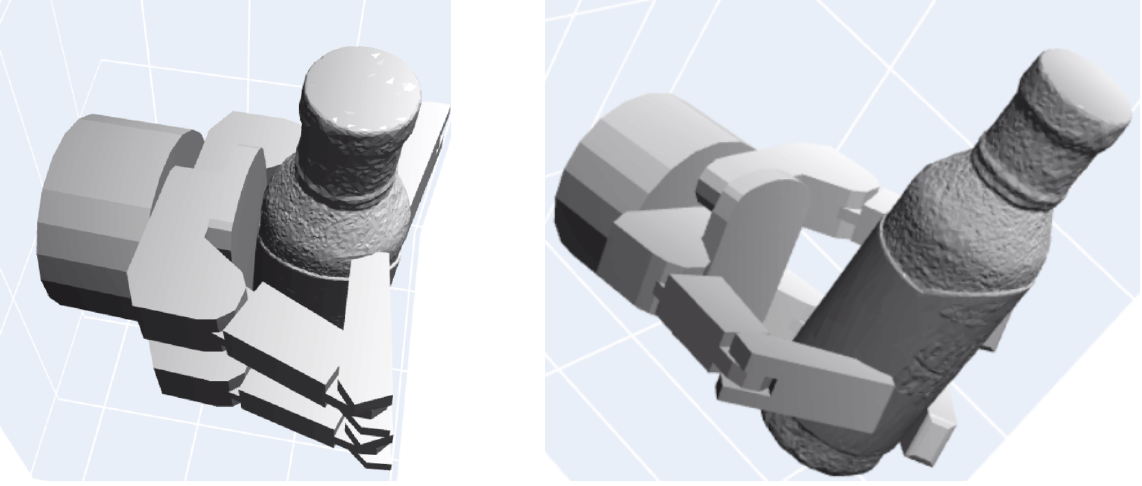}
    \caption{Comparison of grasping strategies on the Barrett hand. 
    Left: a form-closure grasp favored by DRO, where the wrist pose is close to the object and fingers wrap around it. 
    Right: a force-closure grasp favored by MachaGrasp with KAL, where fingertip contacts dominate.}
    \label{fig:barrett_grasp_modes}
\end{figure}

\noindent\textbf{Discussion.} While in the DRO setup, MachaGrasp improves grasp success for ShadowHand (+14.7\%) and Allegro (+5.3\%), we observe a 6.3\% drop for the Barrett hand. Visualization indicates that DRO-generated Barrett grasps typically succeed through \emph{form-closure}, achieved when the wrist pose is close to the object. In contrast, MachaGrasp, guided by the Kinematic-Aware Articulation Loss (KAL), emphasizes fingertip-relevant motions and thus favors wrist poses farther from the object, enabling \emph{force-closure} grasps (see Fig.~\ref{fig:barrett_grasp_modes}). This does not reflect a drawback of KAL itself, but rather a shift in preferred grasping strategy. For hands with limited DoFs such as Barrett, this shift may occasionally reduce compatibility with DRO’s wrist pose proposals, whereas for more dexterous hands (ShadowHand, Allegro) it leads to notable gains.  

\subsection{Few-Shot Generalization to Unseen Hand}
\label{sec:exp-fewshot}

To evaluate the ability of MachaGrasp to generalize across embodiments, we conduct a few-shot adaptation experiment on an unseen hand: the \textbf{Robotiq 3-Finger} (11 DoF), addressing Q3.  
We adopt the Robotiq3F dataset from MultiGripperGrasp
dataset~\cite{casas2024multigrippergrasp} and randomly sample \textbf{100 objects}, each with \textbf{10 grasp poses}, as the training set for this new hand. The model is then fine-tuned with these limited examples and evaluated in Isaac Gym using the predicted wrist poses (following the same mapping procedure as in Sec.~\ref{sec:exp-graspnet}).  

For evaluation, we use our unseen object test set. The simulation results show that MachaGrasp achieves a grasp success rate of \textbf{85.6\%}, demonstrating strong few-shot generalization to previously unseen hands.


\subsection{Real-Robot Experiments}
\label{sec:real-robot}

To further validate MachaGrasp in the physical world (Q4), we deploy the \textbf{Robotiq 3-Finger} hand mounted on a Franka Panda arm.
Object point clouds are captured using three Intel RealSense D435 cameras placed around the workspace.
For each test object, wrist poses are predicted by 6-DoF GraspNet and mapped to the hand-specific wrist frame, after which MachaGrasp (fine-tuned on the Robotiq 3F in Sec.~\ref{sec:exp-fewshot}) predicts the articulation. The wrist pose and articulation are then executed to determine grasp validity.

We evaluate grasp performance on a set of \textbf{10 previously unseen objects}, with 10 grasp attempts per object. Candidate poses are filtered based on Robotiq 3-Finger DoF constraints and physical feasibility (e.g., table collisions). Example predicted grasps on the test objects are illustrated in Fig.~\ref{fig:real_exp}. The system achieves a grasp success rate of \textbf{87\%}, demonstrating that MachaGrasp transfers effectively from simulation to the real world.
\section{CONCLUSION}
We presented MachaGrasp, an eigengrasp-based framework for cross-embodiment dexterous grasp generation. MachaGrasp converts each hand’s URDF into structured morphological tokens to obtain a morphology embedding and hand-specific eigengrasps, and supervises articulation prediction with a Kinematic-Aware Articulation Loss that captures morphology-dependent kinematics beyond joint-space errors. Extensive experiments in simulation and on real hardware demonstrate strong generalization to novel objects and effective transfer across multiple dexterous hands.

Future work will extend this single-shot formulation to trajectory-level generation, enabling the asynchronous and phase-shifted finger coordination required for contact-sensitive, obstacle-aware closing behaviors. Another promising direction is to move from the current modular pipeline to tighter coupling between wrist pose and articulation.

\bibliographystyle{IEEEtran}
\bibliography{refs}

\end{document}